\ifdefined\XeTeXversion\else\pdfoutput=1\fi
\documentclass[11pt,a4paper]{article}
\usepackage{iftex}
\ifPDFTeX\usepackage[utf8]{inputenc}\fi
\usepackage[T1]{fontenc}
\usepackage{lmodern}
\usepackage[margin=25mm]{geometry}
\usepackage{amsmath,amssymb,graphicx,booktabs,tabularx,array}
\usepackage{microtype}
\usepackage[numbers,sort&compress]{natbib}
\usepackage[font=small,labelfont=bf]{caption}
\usepackage[unicode,colorlinks=true,linkcolor=black,citecolor=black,urlcolor=blue]{hyperref}
\usepackage{xurl}
\hypersetup{pdftitle={DRVEXO-4B-Unified: Design and Evaluation of a Modular Speech--Vision System},pdfauthor={Zeyuan Li (Yujie Intelligence)}}
\newcolumntype{Y}{>{\raggedright\arraybackslash}X}
\newcommand{\model}{DRVEXO-4B-Unified}

\newcommand{\ThinkingMMMUCorrect}{668}
\newcommand{\ThinkingMMMUScore}{74.22}
\newcommand{\ThinkingMMMUFreshCount}{445}
\newcommand{\ThinkingMMMUFreshTokens}{3,552,889}
\newcommand{\ThinkingMMMUSeconds}{12,828.77}
\newcommand{\ThinkingMMMURate}{276.95}

\newcommand{\SubsetMMLUCorrect}{2338}
\newcommand{\SubsetMMLUScore}{77.93}
\newcommand{\SubsetMMLUCount}{3000}

\title{\textbf{DRVEXO-4B-Unified:}\\Design and Evaluation of a Modular\\Speech--Vision System}
\author{Zeyuan Li\\\small Yujie Intelligence\\\small Hangzhou, China\\\small \texttt{lizeyuan@drvexo.com}}
\date{November 7, 2026}

\begin{document}
\maketitle
\begin{abstract}
We present \model, a modular multimodal assistant built around a 4B language backbone. A self-contained checkpoint combines text, image, and speech inputs with reference-conditioned speech output and structured interaction. A temporal residual connector adapts acoustic sequences to the language interface; explicit text boundaries support incremental speech synthesis; and three task-conditioned scoring heads rank supplied action candidates alongside generative tool calls. These interfaces connect conversation to application-defined choices while leaving action authorization to the application. The system obtains \ThinkingMMMUScore\% on all 900 MMMU validation questions, exceeding the published validation scores of Gemma 3 4B IT and InternVL3.5-4B without test-time scaling. Strict prompt-level IFEval accuracy is 82.26\% and MATH-500 accuracy is 89.40\%, above the original Qwen3-4B non-thinking report's 81.2\% and 84.8\%, respectively. On NVIDIA GB10, the model achieves a speech-synthesis first-chunk latency of 390\,ms. For concurrent text generation, a growth-aware admission policy completes a 128-answer batch at 594.41 output tokens/s, including filling and draining, without preemptions. We plan to release the model weights and accompanying code under Apache License 2.0, retaining applicable third-party terms.
\end{abstract}

\section{Introduction}
An assistant for a physical application must connect communication with a constrained action space. A user may speak about an image, refer to an earlier turn, or request one of several available actions. These operations require different interfaces: perception produces representations, language generation produces an explanation, speech synthesis produces a waveform, and candidate scoring ranks supplied alternatives. Their integration raises a basic evaluation problem. Successful loading, a correct textual answer, intelligible speech, and a permissible device action are separate outcomes; one cannot stand in for the others.

\model\ is developed by Yujie Intelligence for research on conversational interfaces to assistive-exoskeleton applications. It combines Qwen3.5-4B visual-language components, Qwen3-ASR acoustic components, and Qwen3-TTS speech generation with a temporal connector and task-specific scoring heads~\citep{Qwen2026Model,Qwen2026ASR,Qwen2026TTSPaper}. The label ``4B'' denotes the language backbone. The complete checkpoint stores approximately 6.97 billion scalar tensor elements across its eight components, including speaker state and other buffers; this storage count is not a deduplicated trainable-parameter count. Comparisons with models having approximately 3--4B-class language backbones therefore concern backbone scale, rather than equal total system size. The target application motivates the interfaces, but this report does not evaluate a clinical intervention or autonomous device policy.

Qwen3-Omni studies multimodal understanding and generation through a jointly trained architecture, with separate task evaluations and a controlled study of modality retention~\citep{Xu2025Omni}. Composing pretrained components raises a related question: which capabilities can be established for the assembled system, and under which measurement conditions? In \model, inherited foundations supply the visual-language and speech components, while generated text conditions the active speech-output route. This modular setting calls for explicit interface specifications and evaluation of the resulting system, including controls that distinguish inherited capability from local adaptation.

\begin{figure}[tb]
\centering
\includegraphics[width=\linewidth]{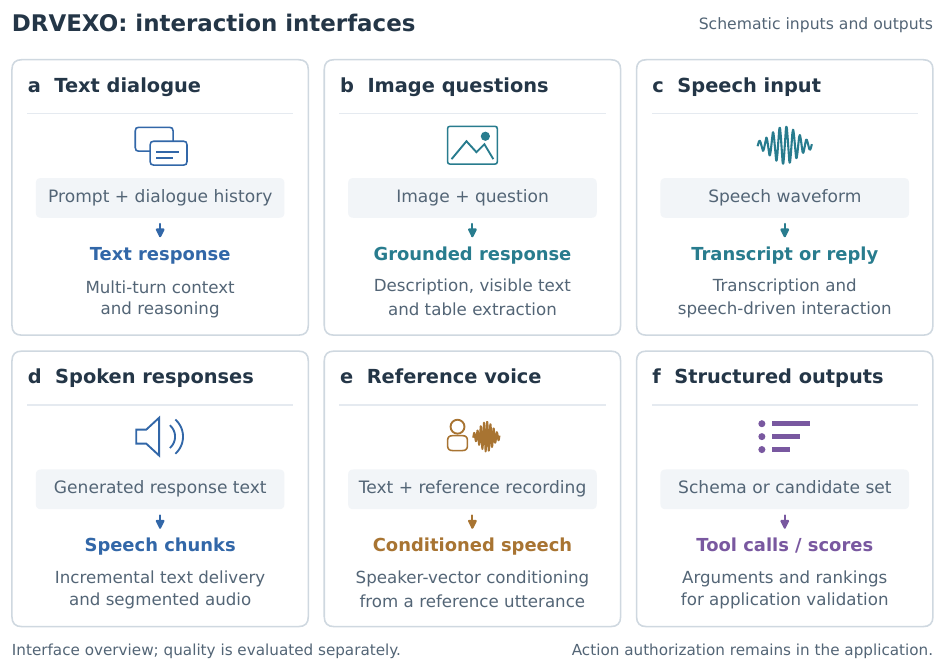}
\caption{Interaction interfaces in \model. Each panel identifies an input and its corresponding output. Reference-voice conditioning modifies speech output; structured outputs provide tool arguments or candidate scores for application-side validation. The examples are schematic interface descriptions, not measured conversations. Action authorization remains outside the model.}
\label{fig:capabilities}
\end{figure}

Figure~\ref{fig:capabilities} summarizes the interaction interfaces. Text, images, and speech enter through their respective input paths; responses can be returned as text, segmented speech, or structured outputs. Reference recordings provide speaker conditioning, while candidate sets define the alternatives to be ranked. These interfaces allow an application to combine conversation with explicit action choices without treating a generated explanation as an execution command.

The system combines three interaction choices: temporally resolved speech input, generated-text boundaries for streaming speech and reference-voice conditioning, and separate interfaces for free-form tool generation and supplied-candidate ranking. The report examines these choices through component specifications, public task results, comparison with published model reports, and measurements of inference capacity and reference-runtime speech delivery. Section~\ref{sec:architecture} describes the model, and Section~\ref{sec:training} presents training and checkpoint assembly. Section~\ref{sec:evaluation} defines the experimental questions and measurement protocols. Section~\ref{sec:results} analyzes the results, followed by the limitations in Section~\ref{sec:limitations}. We plan to release the model and accompanying code under Apache License 2.0, with upstream attribution and applicable third-party terms retained.

\section{Model Design}
\label{sec:architecture}
The system uses explicit interfaces between perception, language, speech generation, and application decisions. Acoustic features pass through a trainable temporal connector before entering the language space. Images use the inherited visual projection. Generated text forms the boundary between language reasoning and speech synthesis, allowing the same response to be inspected, delivered as text, or segmented for speech. Candidate scores form a separate output interface because ranking supplied actions and generating an explanation have different supervision and validation requirements.

\subsection{Components and checkpoint representation}
The release uses a shared safetensors index with namespaced tensors. The inspected checkpoint contains 16 weight shards and 2,131 tensors distributed across eight components (Table~\ref{tab:components}). A strict reader resolves the component namespace and checks tensor availability. This organization supports component-wise loading and explicit precision accounting; it does not imply that all components share parameters or that an unmodified generic model loader can execute every task.

\begin{table}[htbp]
\centering\small
\caption{Checkpoint composition. Elements are scalar entries counted from saved tensor shapes, in millions (M), including non-trainable state. Counts do not deduplicate tied parameters. The BF16 release preserves source FP32 parameters where applicable; the release label does not imply uniform precision for every tensor or operator.}
\label{tab:components}
\begin{tabularx}{\linewidth}{@{}lrrY@{}}
\toprule
Component & Tensors & Elements (M) & Function\\
\midrule
Language backbone & 426 & 4,205.751 & Contextual representations and text generation\\
Vision encoder & 297 & 333.514 & Image encoding and projection\\
Audio encoder & 397 & 317.478 & Acoustic feature extraction\\
Connectors & 28 & 14.580 & Temporal adaptation and projection\\
Speech generator & 480 & 1,928.677 & Speech-code generation\\
Speech codec & 496 & 170.557 & Code and waveform conversion\\
Speaker representation & 1 & 0.002 & Speaker-conditioning state\\
Candidate decision heads & 6 & 0.008 & Three task-specific linear scorers\\
\midrule
Total & 2,131 & 6,970.568 & Eight components, 16 weight shards\\
\bottomrule
\end{tabularx}
\end{table}

\begin{figure}[tb]
\centering
\includegraphics[width=\linewidth]{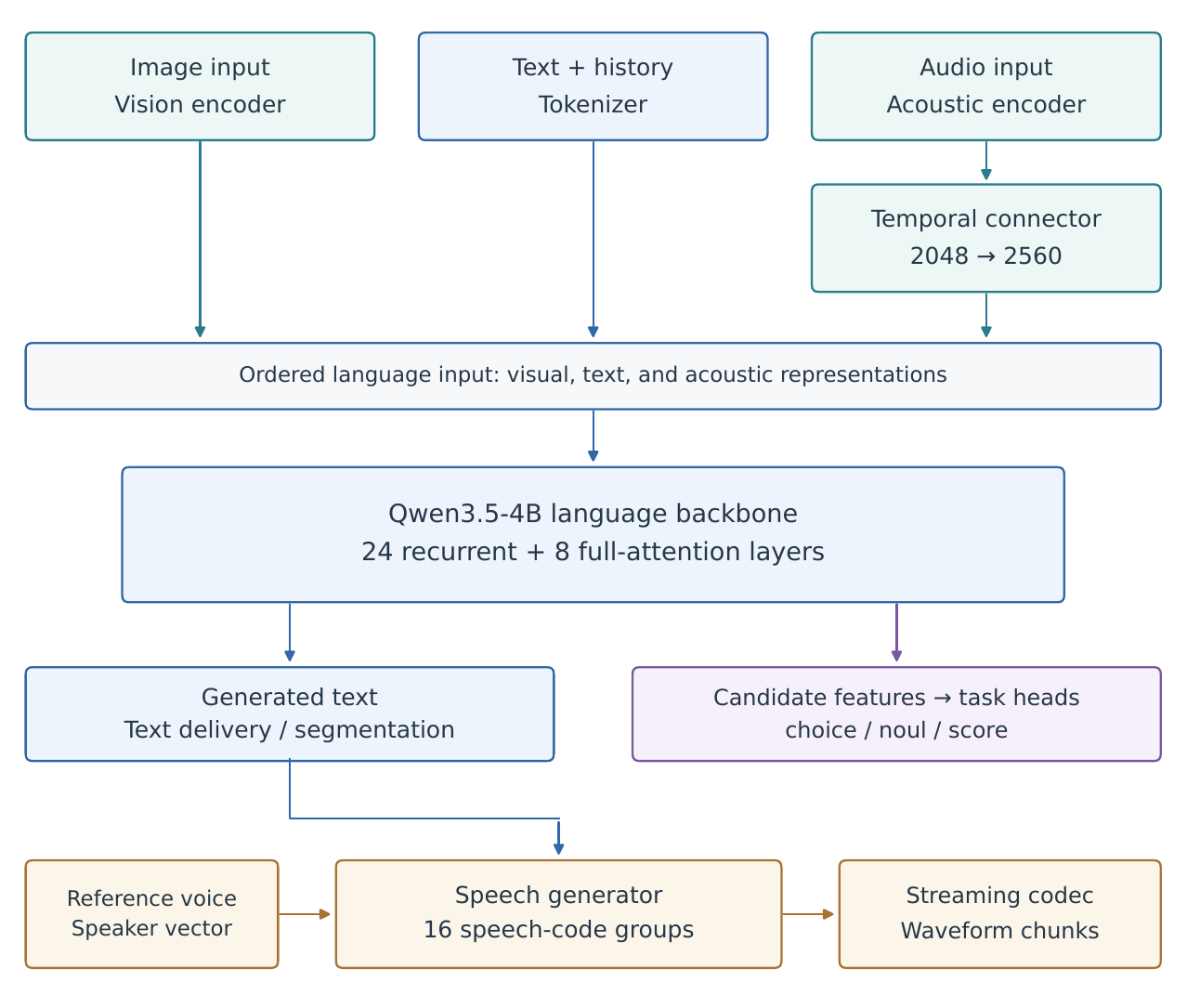}
\caption{Implemented component interfaces in \model. Speech output is conditioned on generated text and an optional reference voice. Candidate scores are returned to the application; action authorization and execution remain outside the model. The diagram describes software and checkpoint composition, rather than a claim of joint end-to-end pretraining.}
\label{fig:architecture}
\end{figure}

Figure~\ref{fig:architecture} summarizes the implemented paths. Image and speech inputs are encoded into representations consumed by the language backbone. Text and committed conversation history enter the same language path. Generated text can be delivered directly or segmented for speech synthesis. Candidate heads score supplied alternatives, and a reference recording supplies speaker conditioning for speech output.

\subsection{Visual-language backbone}
The language, vision, and tokenizer components originate from Qwen3.5-4B~\citep{Qwen2026Model}. The language backbone has 32 layers, hidden width 2,560, and intermediate width 9,216. Its hybrid attention layout contains 24 linear-attention layers and eight full-attention layers, with one full-attention layer following each group of three linear-attention layers. The configuration specifies 16 attention heads and four key/value heads. The vision encoder has 24 layers, hidden width 1,024, output width 2,560, and patch size 16.

The vision projection produces representations at the language backbone's width, allowing image content and the user's question to be processed in the same conversational context. The processor supplies image patches and positional information; the language decoder then generates the textual response. The interface supports image description, questions about visible text, and extraction from charts or tables through the same generation path.

\subsection{Temporal speech connector}
The inherited speech-recognition component belongs to the Qwen3-ASR family~\citep{Qwen2026ASR}. Its retained audio configuration uses a 24-layer encoder with model width 1,024, 128 mel bins, and 16 attention heads. The temporal connector operates on the extracted 2,048-dimensional features; the connector input dimension and the encoder's internal model width refer to different interfaces. Upstream speech-recognition results are not measurements of this composed checkpoint.

Let $H\in\mathbb{R}^{T\times2048}$ denote the connector input. A layer-normalized bottleneck maps each frame to width 256:
\begin{equation}
 U_0 = \operatorname{GELU}\!\left(\operatorname{LN}(H)W_{\mathrm{down}}+b_{\mathrm{down}}\right).
\end{equation}
Two depthwise temporal convolutions with kernel size five use dilations one and two, respectively. Each is followed by feature mixing and a residual connection. Padding preserves the frame count $T$. An output projection returns the bottleneck representation to width 2,048:
\begin{equation}
 \widetilde H=H+U_2W_{\mathrm{up}}+b_{\mathrm{up}}.
\end{equation}
Here $\operatorname{LN}$ denotes layer normalization and $\operatorname{GELU}$ denotes the Gaussian error linear unit.
The output branch is zero-initialized in the connector definition, giving an identity residual at initialization; this is an initialization property rather than a claim that the final learned branch remains zero. A separate layer-normalized multilayer projection maps the result into the 2,560-dimensional language space. A connectionist temporal classification (CTC) readout is used in the training implementation and does not act as a semantic routing gate at inference.

Preserving the frame sequence allows the language projection to consume temporally resolved acoustic features. The dilated residual blocks expand the temporal receptive field while the bottleneck limits the width of the trainable branch. The acoustic encoder, temporal adaptation, and language projection therefore have distinct roles in the speech-input path.

\subsection{Speech generation and custom voices}
Speech output is conditioned on the response text. The speech generator converts text into sequences of acoustic codes, and the codec reconstructs waveform segments from those codes. The generator has 28 layers with hidden width 2,048 and uses 16 speech-code groups. Keeping generated text as an explicit intermediate representation allows the application to present the same response in written and spoken form and to schedule synthesis as text segments become available.

Custom voice conditioning supplies an x-vector extracted from a reference recording to the speech generator. The interface accepts finite mono recordings of one to 30 seconds and does not require per-speaker retraining. This separates the linguistic content, specified by the response text, from the speaker condition, specified by the reference utterance. The conditioning vector is attached to the synthesis request so that separate conversations can use different references.

For incremental delivery, the text segmenter releases sentence-sized units at sentence boundaries, eligible soft separators, or a maximum buffer of 48 characters. Decimal points are distinguished from sentence endings. The codec reconstructs successive waveform chunks using retained left context and discards samples already delivered. Chunk size and context length are runtime settings; the speech experiments distinguish native and reference configurations. Bounded queues and cooperative cancellation coordinate text generation, synthesis, and playback.

\subsection{Candidate Decisions and Structured Interaction}
\subsubsection{Task-conditioned candidate scoring}
The decision interface accepts two to 151 supplied candidates and a task identifier that selects one of three task-specific scorers. For a candidate representation $h_i\in\mathbb{R}^{2560}$ and task $t$, the corresponding linear head computes
\begin{equation}
 z_i=w_t^\top h_i+b_t,\qquad \hat\imath=\arg\max_i z_i.
\end{equation}
The heads contain three weight tensors and three bias tensors. Their outputs are relative candidate scores, not calibrated probabilities of physical success. Candidate sets and their interpretation are application inputs; selecting a candidate does not directly actuate a device.

The task identifier selects the scoring head while the supplied candidate set determines the available alternatives. Returning both scores and the selected index allows an application to inspect the ranking and apply its own acceptance policy. The head does not create an alternative outside the supplied set.

\subsubsection{Tool calls, schemas, and execution boundaries}
The application interface supports tool calls and JavaScript Object Notation (JSON) Schema outputs. A request supplies the available tool definitions or the required response schema; the language model generates the associated name, arguments, or structured fields. Parsing and schema validation check the representation before it is passed to the application. Candidate scoring complements this generative interface when the alternatives are already enumerated.

Open-ended dialogue is interpreted by the language model using committed conversation history. It is not intercepted by keyword lists or fixed identity replies. Deterministic checks instead govern authentication, wire formats, bounds, request cancellation, duplicate protection, and action authorization. This separation permits natural references in conversation while keeping actual device execution subject to an application control gate.

\section{Training and Checkpoint Assembly}
\label{sec:training}
\subsection{Inherited components and bounded component training}
Training is component-specific. The visual-language backbone and acoustic encoder inherit their foundation-model training. Speech generation and codec components originate from Qwen3-TTS-12Hz-1.7B-Base and its speech stack~\citep{Qwen2026TTS,Qwen2026TTSPaper}. Local component training addresses the acoustic--language interface and candidate scoring. Table~\ref{tab:training} separates these optimization stages from checkpoint assembly. No new joint foundation-model pretraining stage is introduced.

\begin{table}[htbp]
\centering\small
\caption{Training and assembly boundaries. Recipe information is distinguished from measured quality of the assembled release. The component experiments are not a controlled ablation of the complete system.}
\label{tab:training}
\begin{tabularx}{\linewidth}{@{}lYY@{}}
\toprule
Stage & Optimization / frozen components & Role in the system\\
\midrule
Inherited foundations & Upstream language, vision, acoustic, and speech objectives & Perception, language generation, and speech synthesis\\
Temporal connector & Answer-token likelihood plus 0.1 CTC; language backbone frozen & Temporally resolved acoustic--language adaptation\\
Candidate heads & Candidate-distribution cross-entropy; task-specific linear scorers & Ranking supplied alternatives for each task\\
Checkpoint assembly & Namespace consolidation and precision restoration; no optimizer step & A self-contained checkpoint with explicit component loading\\
\bottomrule
\end{tabularx}
\end{table}

The retained temporal-connector training implementation keeps the language backbone frozen and combines answer-token negative log likelihood with an auxiliary CTC objective:
\begin{equation}
 \mathcal{L}_{\mathrm{speech}}=-\sum_{t\in\mathcal{A}}\log p_\theta(y_t\mid y_{<t},\widetilde H)+0.1\,\mathcal{L}_{\mathrm{CTC}},
\end{equation}
where $\mathcal{A}$ contains answer positions rather than prompt or padded positions. The implementation uses FP32 optimizer states and CTC loss with a differentiable BF16 serving forward. The auxiliary CTC readout supports alignment training and is not an inference-time intent gate. This source-level recipe is not an ablation showing a gain for the released connector, nor a fresh speech-quality score for the complete checkpoint.

The retained bounded connector recipe specifies 2,048 steps, batch size four, and AdamW optimization. Peak learning rates are $10^{-5}$ for the base projection, $10^{-4}$ for the temporal branch, and $3\times10^{-4}$ for the auxiliary CTC readout. This description records the historical recipe without promoting its development-set observations to an official speech benchmark. Establishing connector utility requires a paired evaluation with the same acoustic inputs and backbone, together with a documented mapping from the evaluated connector to the release.

\subsection{Candidate-head supervision}
For each supplied candidate set, head training minimizes cross-entropy between a target distribution and the softmax over candidate scores. Checkpoint selection uses equal-task macro cross-entropy across \texttt{choice}, \texttt{noul}, and \texttt{score}, giving each task equal weight despite different sample counts. The component experiment uses its paired frozen reference backbone. Section~\ref{sec:decision-results} reports that validation separately from evaluation with the assembled release.

\subsection{Assembly and precision verification}
Assembly resolves each component namespace into a shared safetensors index. Tensor counts, shapes, source bindings, and precision-restoration receipts provide loading evidence. They do not prove end-to-end conversational quality. For native evaluation, 723 language and vision tensors are copied into a backend-compatible view and checked against the release; 48 original FP32 language parameters are restored exactly after loading. The view serves evaluation and does not replace the complete eight-component package.

\section{Experimental Design}
\label{sec:evaluation}
\subsection{Evaluation questions}
The experiments separate task correctness from interface implementation and inference capacity. We first test whether the assembled system can answer knowledge and reasoning questions, follow explicit instructions, and extract information from images, then place these results alongside published evaluations of related models. We examine the two structured-output interfaces: ranking supplied alternatives and generating tool calls. We also inspect archived speech-delivery traces to distinguish response startup from waveform production, and measure whether long responses can finish under concurrent execution without exhausting the key/value (KV) cache. Table~\ref{tab:scope} links each question to the corresponding task and metric.

Two design choices are central to the study. First, general task scores and application-specific decisions are evaluated separately because they exercise different outputs and scoring rules. Second, restricted generation and long-thinking generation are treated as distinct protocols. Saved-answer diagnostics examine how output limits and extraction failures affect interpretation; they are not used to infer a component's causal contribution. Speech quality and physical-device outcomes require additional measurements beyond these text, image, and decision tasks.

\begin{table}[htbp]
\centering\small
\caption{Experimental questions and measurements. Sample counts define evaluation scope; measured outcomes are reported in Section~\ref{sec:results}.}
\label{tab:scope}
\begin{tabularx}{\linewidth}{@{}YYY@{}}
\toprule
Question & Task and scope & Measurement\\
\midrule
Knowledge and reasoning across domains & MMLU-Pro~\citep{Wang2024MMLUPro}: 12,032 restricted-protocol questions; 3,000 stratified long-thinking questions & Answer accuracy and subject-level variation\\
Instruction and mathematical constraints & IFEval~\citep{Zhou2023IFEval}: 541 prompts; MATH-500~\citep{H42024Math}: 500 problems & Strict/loose compliance; verified final answers\\
Image-grounded reasoning & MMMU~\citep{Yue2024MMMU}: 900 validation questions; ChartQA~\citep{Masry2022ChartQA}: 1,920 validation questions & Overall, subject-level, and question-source accuracy\\
Structured decisions and tool use & JevBench~\citep{BenchmarkHeaven2026JevBench}: 231 public examples; BFCL~\citep{Patil2025BFCL}: 3,641 single-turn cases & Candidate accuracy; category-specific function-call correctness\\
Streaming speech delivery & Archived merged-model synthesis and pre-merge reference-runtime requests & First PCM, total response time, audio duration, and replay continuity\\
Capacity during growing responses & Short windows, complete batches, sustained admission & Output tokens/s, cache occupancy, preemptions, output completion\\
\bottomrule
\end{tabularx}
\end{table}

\subsection{Tasks, samples, and metrics}
\paragraph{Text reasoning and instruction following.}
MMLU-Pro~\citep{Wang2024MMLUPro} supplies 12,032 questions across 14 subjects and official prompt-formatting and answer-extraction helpers. The restricted protocol evaluates the full set. To bound the cost of long-thinking evaluation while preserving subject coverage, we select a fixed 3,000-question proportional stratified subset. Subject quotas use the largest-remainder method; within each subject, a fixed seeded hash of question IDs determines selection. Neither answers nor model correctness enter the selection rule. The subset score is therefore reported with its own denominator, separately from the full-set score.

IFEval~\citep{Zhou2023IFEval} evaluates compliance with verifiable instructions using all 541 official prompts and the official strict and loose checkers. We report both prompt-level compliance, which requires all instructions in a prompt to pass, and instruction-level compliance. MATH-500~\citep{H42024Math} is a 500-problem subset of the MATH test set~\citep{Hendrycks2021Math}; we evaluate all 500 problems. Prompts request a step-by-step solution ending in a boxed answer; Math-Verify 0.9.0~\citep{HuggingFace2025MathVerify} checks the final answer. These tasks distinguish formatting constraints from mathematical answer correctness.

\paragraph{Visual understanding.}
MMMU~\citep{Yue2024MMMU} tests image-grounded reasoning using all 900 public validation questions, with 30 examples in each of 30 subjects. We use the official response parser and deterministic per-ID fallback and retain subject-level counts. ChartQA~\citep{Masry2022ChartQA} tests chart reading and associated reasoning on the available 1,920-example public validation split, evenly divided between human-authored and machine-generated questions. Scoring uses the relaxed answer criterion, and both groups are reported to reveal differences hidden by the aggregate. These are public-validation measurements, not hidden-test leaderboard scores.

\paragraph{Candidate decisions and function calls.}
JevBench~\citep{BenchmarkHeaven2026JevBench} evaluates typed decisions on 231 public examples. Features are generated by the assembled language backbone and scored by the retained FP32 candidate heads. We compare normalized states against the checked training states to detect exact overlap; this check does not establish semantic independence or absence from foundation pretraining. The public-subset accuracy is distinct from the sealed-board composite score. We also report the historical head-validation experiment with its paired frozen reference backbone to examine variation among the three scoring tasks; its 3,202 examples use the recorded target-tie accuracy criterion and cross-entropy.

The Berkeley Function Calling Leaderboard (BFCL)~\citep{Patil2025BFCL} tests generated function selections and arguments. Our scope comprises 3,641 single-turn cases in 13 categories, including simple, multiple, parallel, relevance, and irrelevance cases. Multi-turn and agentic categories are excluded. Category-specific checkers assess function-call correctness, while missing final responses remain in the denominator. The resulting scope must not be labeled the complete BFCL v4 overall score. Appendix~\ref{app:bfcl} describes the absent-tool normalization required for a uniform saved-answer rescore.

\subsection{Generation protocols and answer preservation}
\label{sec:protocols}
The restricted protocol uses greedy decoding and an 8,192-token context. IFEval and MATH-500 use a non-thinking chat template with output budgets of 2,048 and 4,096 tokens, respectively. The MATH prompt still requests explicit reasoning in the answer text. Restricted MMMU and ChartQA generation allows 1,024 new tokens and uses a declared image-pixel budget of 1,048,576. These IFEval, MATH-500, and ChartQA measurements are retained without a long-thinking rerun.

Restricted MMLU-Pro uses a raw completion prompt with up to five same-subject chain-of-thought exemplars. An input-budget rule removes exemplars when the formatted prompt reaches 2,048 tokens. Generation allows 2,048 new tokens and stops at the configured next-question delimiter; unparseable answers count as incorrect. This length-adapted setup differs from an unmodified five-shot prompt.

The long-thinking protocol keeps all five MMLU-Pro exemplars, uses a 131,072-token context, and allows 32,768 new tokens for MMMU, MMLU-Pro, and BFCL. Sampling uses temperature 1.0, top-$p$ 0.95, top-$k$ 20, minimum-$p$ 0, and presence penalty 1.5. Visual inputs retain original images and source-processor defaults. Only the final response after the thinking delimiter is scored. A fixed per-ID seed defines one response per question; completed answers are not regenerated to select a better response. The settings define our experiment and do not establish exact parity with a vendor's prompts, repetitions, or leaderboard aggregation.

For MMLU-Pro, 624 completed answers selected by the frozen subset manifest are reused byte-for-byte and the remaining 2,376 are generated once. Resumed MMMU stages follow the same preservation rule. Retained answers contribute to capability denominators, but their previously generated tokens and elapsed time are excluded from a later stage's throughput. Output limits, missing final delimiters, and unparseable answers are recorded alongside accuracy rather than removed from the scored set.

\subsection{Comparison with published model evaluations}
\label{sec:comparison-method}
We use author-published technical reports and official model cards to place the results in context, with Qwen3.5-4B as the direct visual-language foundation~\citep{Qwen2026Model}. Other comparison models have a similar language-backbone scale. An active-parameter count for a mixture-of-experts model is not treated as the total size of a dense model. None of the comparison models is rerun in this study.

For each source, we record the model version, task, available split information, and evaluation mode. Tables identify stratified and full samples, prompt-level instruction compliance, and the included function-calling categories. Missing source details are marked as unspecified. Comparisons are presented by task, without constructing a composite rank.

\subsection{Speech-delivery measurements}
\label{sec:speech-method}
Speech startup and complete-response timing answer different questions from aggregate text throughput. We inspect existing local HTTP traces on NVIDIA GB10. For the merged checkpoint, a short fixed-text synthesis request uses the three-stage native backend after earlier synthesis and visual requests have warmed the service. For longer conversational traces, a separate pre-merge reference runtime uses a text prompt or a prerecorded synthetic utterance paired with an image. These two runtime configurations are reported separately. Each selected scenario has one recorded observation; there is no repeated-trial latency distribution or controlled cold-start comparison.

Timing begins immediately before the client issues the request. First pulse-code modulation (PCM) latency ends when the client receives its first audio event; completion ends at the final event for native synthesis and at HTTP-stream closure for the reference runtime. Audio duration is the number of waveform samples divided by sample rate. We define the request-level real-time factor (RTF) as completion time divided by output-audio duration. Inter-chunk gaps and replayed playback-buffer events describe delivery continuity. The synthetic-audio-and-image case includes the request's recognition, image processing, response generation, and synthesis, but excludes recording time and input creation. None of these traces measures the interval from a physical speaker's last utterance to audible loudspeaker playback, or latency over a remote network.

\subsection{Inference experiment design}
\label{sec:systems-design}
All native text and visual inference measurements use NVIDIA GB10 and vLLM 0.33. The main backbone runs in BF16, with its 48 source FP32 language parameters restored after loading. JevBench uses the reference backend and is excluded from native-engine throughput comparisons. We report the execution path separately because native language generation and complete spoken interaction exercise different components.

\paragraph{Why measure both windows and complete responses?}
Fixed windows measure how rapidly an engine produces tokens while many requests are active. They may end before the longest requests consume their full cache or finish their answers. We therefore use three complementary workloads: a concurrency sweep with 30-second warmup and 90-second measurement windows; fixed batches that drain to natural completion; and sustained admission with cache and preemption telemetry. In the short sweep, only the probe's own unfinished requests are canceled at a boundary. In complete-response experiments, requests retain their original generation budgets and finished answers remain saved.

Aggregate output throughput is
\begin{equation}
 R_{\mathrm{agg}}=\frac{\sum_{j=1}^{N}n_j}{t_{\mathrm{generation}}},
\end{equation}
where $n_j$ is the generated output length and the interval includes prompt processing, scheduling, output handling, interleaved checking, and the final response tail. Initial model loading and separate post-generation rescoring are excluded. Complete-batch and complete-stage measurements use saved outputs; fixed windows include in-flight tokens and are labeled separately. We additionally record actual running requests, KV occupancy, preemptions, length-limited outputs, and missing final responses.

\paragraph{Admission under context growth.}
Dynamic cache allocation and sharing reduce redundant storage~\citep{Kwon2023PagedAttention}, but admitted requests continue to grow after new intake pauses. The evaluated admission policy therefore combines observed allocation with estimated future demand. Let $B$ be the physical pool size, $A$ its observed allocated blocks, $C$ the lease for newly admitted prefills, $F$ the further-growth and scratch estimate for acknowledged requests, and $G$ a publication guard. The estimate is
\begin{equation}
 \widehat B=A+C+F+G.
\end{equation}
Shared physical allocation is counted once in $A$. A prefill remains charged for its prompt and a forecast output window until it has streamed output and is at least ten seconds old. An acknowledged request leases 1,024 further tokens, bounded by its unchanged remaining budget, plus recurrent-state scratch and one publication block. The global guard is 32 blocks. These leases are refreshed as outputs grow and do not reserve every possible remaining token.

The policy uses $B=2{,}048$ blocks, a working level of $\lfloor0.85B\rfloor=1{,}740$, and a usable level of $\lfloor0.90B\rfloor=1{,}843$. KV capacity is 36\,GiB, the CUDA allocator cap is 48\,GiB, and the process-memory limit is 90\,GiB. The request target starts at 64 and can rise to 128. Intake pauses at observed KV occupancy of 75\%, stale telemetry, sustained waiting pressure, or a new preemption; active requests retain their budgets. A fixed initial batch is filled and then drained without replacement admissions to check complete-response behavior before evaluating continuous refill. Memory limits share one physical-memory platform, and the growth estimate is not a worst-case memory guarantee.

\section{Results and Analysis}
\label{sec:results}
The results address task capability, variation within tasks, and the conditions under which a throughput measurement remains useful. Protocols and sample sets are kept explicit throughout: a score describes its declared workload, while protocol changes and component changes require separate interpretation.

\subsection{Knowledge, reasoning, and instruction following}
Table~\ref{tab:quality} compares knowledge, instruction following, and mathematical reasoning across models at a similar language-backbone scale. It includes the Qwen3.5-4B backbone, both inference modes of the original Qwen3-4B, and Phi-4-mini-reasoning. MMLU-Pro~\citep{Wang2024MMLUPro}, IFEval~\citep{Zhou2023IFEval}, and MATH-500~\citep{H42024Math} occupy separate columns so that each task can be read across models. Complete DRVEXO task counts and protocol labels are retained in Table~\ref{tab:task-details} in Appendix~\ref{app:reproducibility}.

\begin{table}[!htbp]
\centering\small
\caption{Text-task comparison at similar language-backbone scale (\%). Published reference scores are taken from the cited author reports and model cards. IFEval is strict prompt-level accuracy; Phi reports MATH-500 Pass@1. DRVEXO protocols are defined in Section~\ref{sec:protocols}.}
\label{tab:quality}
\begin{tabularx}{\linewidth}{@{}Yrrr@{}}
\toprule
Model and inference mode & MMLU-Pro & \shortstack[r]{IFEval\\(strict prompt)} & MATH-500\\
\midrule
Qwen3.5-4B~\citep{Qwen2026Model} & 79.1 & --- & ---\\
Qwen3-4B, non-thinking~\citep{Yang2025Qwen3,Qwen2025Instruct2507Card} & 58.0 & 81.2 & 84.8\\
Qwen3-4B, thinking~\citep{Yang2025Qwen3,Qwen2025Thinking2507Card} & 70.4 & 81.9 & 97.0\\
Phi-4-mini-reasoning, 3.8B~\citep{Xu2025Phi4Mini} & --- & --- & 94.6\\
\midrule
DRVEXO, restricted (this work) & 59.13 & 82.26 & 89.40\\
DRVEXO, long thinking (this work) & \SubsetMMLUScore\ [a] & --- & ---\\
\bottomrule
\end{tabularx}
\par\smallskip
\begin{minipage}{\linewidth}\small
[a] Fixed 3,000-question, 14-subject stratified subset; completed-run score awaiting independent final rescoring. The restricted MMLU-Pro row uses all 12,032 questions. Qwen3 MMLU-Pro values come from the original-model comparator columns in the later official model cards; its IFEval and MATH-500 values come from Tables 17 and 18 of the original technical report. A dash means that no score for the exact listed metric is available in the cited source or current DRVEXO run.
\end{minipage}
\end{table}

\paragraph{Instruction following and mathematical reasoning.}
DRVEXO's strict prompt-level IFEval score of 82.26\% exceeds the original Qwen3-4B report's non-thinking 81.2\% and thinking 81.9\%, by 1.06 and 0.36 percentage points. On MATH-500, our 89.40\% exceeds the reported Qwen3-4B non-thinking score of 84.8\% by 4.60 points.

The IFEval result comprises 445 of 541 prompts satisfying every required instruction. At the individual-instruction level, 730 of 834 requirements pass strict checking (87.53\%); loose checking gives 90.17\% instruction compliance and 85.77\% prompt compliance. MATH-500 contains 447 correct final answers out of 500. The IFEval and MATH-500 runs contain 12 and 46 length-limited outputs, respectively; these remain in the scored sets, including any that already contain a correct final answer.

\paragraph{Knowledge coverage.}
The completed long-thinking MMLU-Pro run records \SubsetMMLUCorrect/\SubsetMMLUCount\ correct answers (\SubsetMMLUScore\%) on the fixed 14-subject stratified subset. Its 3,000 unique answers include 624 preserved responses and 2,376 fresh responses, with no length-limited or missing-final outputs. Independent final rescoring remains pending, so Table~\ref{tab:quality} identifies this value as the completed-run subset score. The restricted-protocol full-set result is retained as a separate row, with 7,115 correct answers out of 12,032 (59.13\%).

\subsection{Visual reasoning and variation within tasks}
\label{sec:visual-results}
The long-thinking MMMU run obtains \ThinkingMMMUCorrect/900 correct (\ThinkingMMMUScore\%), confirmed by independent CPU rescoring. All 900 public-validation answers are included: 455 are unchanged responses retained from the preceding stage and \ThinkingMMMUFreshCount\ are newly generated. Two length-limited outputs and four outputs without the thinking-close delimiter remain in the denominator. Relative to the restricted protocol, accuracy increases by 15.89 percentage points from 58.33\%, while length-limited outputs decrease from 223 to two. Image handling, the thinking template, sampling, and output budget change together, so the gain cannot be attributed solely to reduced truncation or a single component. Supplementary saved-output diagnostics appear in Appendix~\ref{app:protocol-diagnostics}.

\paragraph{Comparison with published visual-language results.}
Table~\ref{tab:vision-comparison} places this result alongside author-reported scores for small dense visual-language models. Our 74.22\% exceeds Gemma 3 4B IT's 48.8\% and InternVL3.5-4B's 66.6\% on their reported MMMU validation evaluations, by 25.42 and 7.62 percentage points. It also exceeds the reported numbers for Qwen3-VL-4B in both Instruct and Thinking variants and MiniCPM-V 4.0. The table records the split information supplied by each source.

\begin{table}[htbp]
\centering\small
\caption{Author-reported visual-language results and our observations (\%). Models are selected by language-backbone scale; complete-system size is described in Section~\ref{sec:architecture}. ``Unspecified'' marks missing split information in the cited table. Gemma uses Pan \& Scan; InternVL uses no test-time scaling. Dataset splits are listed separately for each benchmark.}
\label{tab:vision-comparison}
\begin{tabularx}{\linewidth}{@{}Yrlrl@{}}
\toprule
Model & MMMU & Split & ChartQA & Split\\
\midrule
Qwen3.5-4B~\citep{Qwen2026Model} & 77.6 & Unspecified & --- & ---\\
Qwen3-VL-4B-Instruct~\citep{Bai2025QwenVL} & 67.4 & Unspecified & 84.6 & Test\\
Qwen3-VL-4B-Thinking~\citep{Bai2025QwenVL} & 70.8 & Unspecified & 88.8 & Test\\
MiniCPM-V 4.0~\citep{OpenBMB2025MiniCPMV4} & 51.2 & Unspecified & 84.4 & Unspecified\\
Gemma 3 4B IT~\citep{Gemma2025Report} & 48.8 & Validation & 68.8 & Test\\
InternVL3.5-4B~\citep{Wang2025InternVL} & 66.6 & Validation & 86.0 & Test\\
\midrule
DRVEXO (this work) & 74.22 & Validation & 81.61 & Validation\\
\bottomrule
\end{tabularx}
\end{table}

\begin{figure}[tb]
\centering
\includegraphics[width=\linewidth]{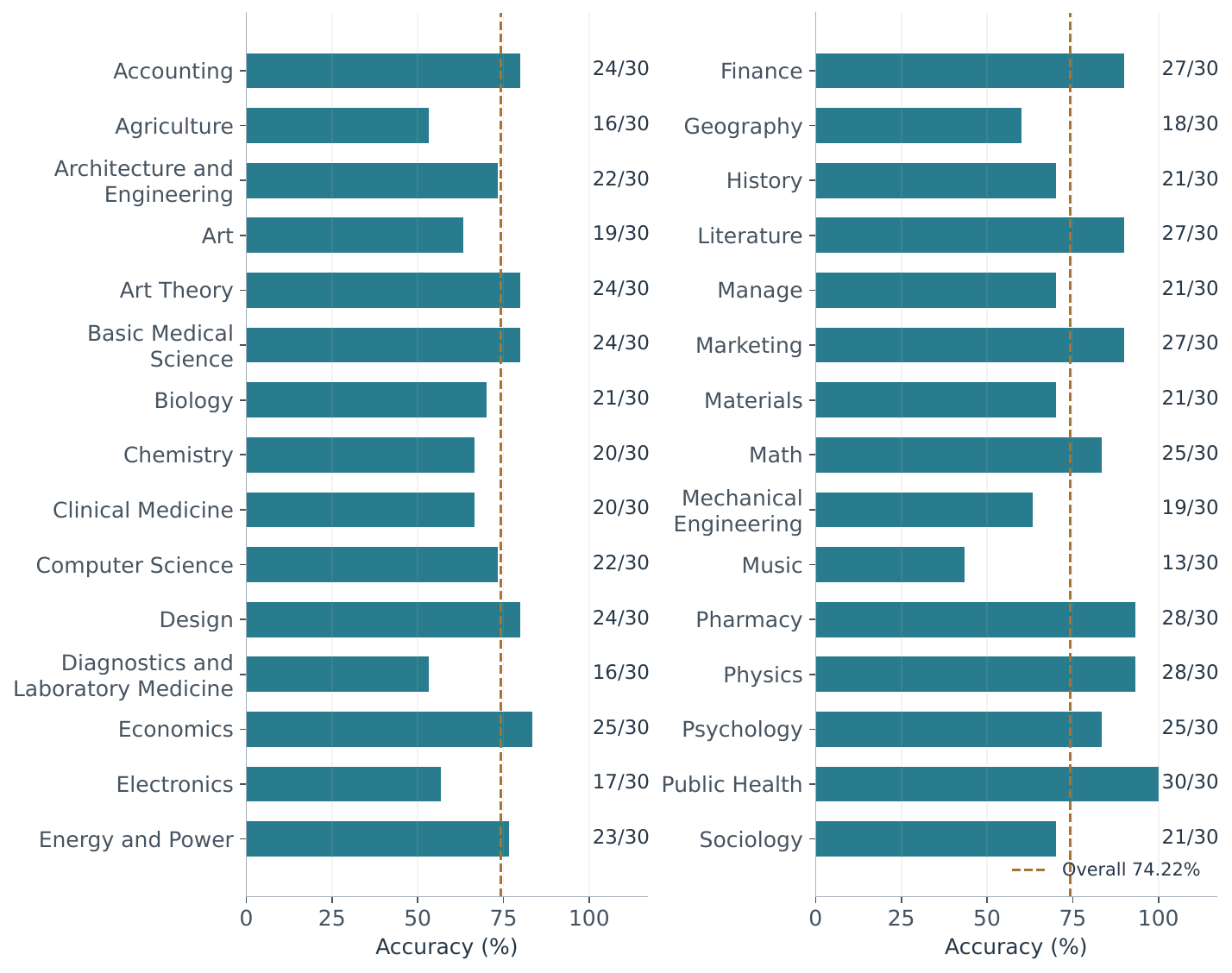}
\caption{Complete long-thinking MMMU validation by subject. All 900 questions are included, with 30 per subject; labels give correct/total counts. The dashed line is the overall 74.22\%. These descriptive single-run results do not identify a causal error mechanism or measure device-task safety.}
\label{fig:subjects}
\end{figure}

\paragraph{Subject-level behavior.}
The aggregate score masks substantial variation (Figure~\ref{fig:subjects}). Each subject contributes 30 questions, giving every subject equal weight. Public Health obtains 30/30, while Music obtains 13/30; Agriculture and Diagnostics and Laboratory Medicine each obtain 16/30. The system therefore remains uneven across the evaluated domains. These descriptive counts identify where closer error analysis is needed, but a subject label does not distinguish failures of visual perception, domain knowledge, or reasoning. Thirty examples per subject also provide limited support for fine-grained rankings.

ChartQA exhibits a similar need for disaggregation. Human-authored questions score 75.00\% (720/960), compared with 88.23\% (847/960) for machine-generated questions. The 13.23-point gap is hidden by the overall 81.61\% score, and neither group contains length-limited outputs. Question source changes several properties of the task, so the gap motivates case-level analysis rather than an attribution to one visual or reasoning operation.

\subsection{Candidate decisions and function calling}
\label{sec:decision-results}
On the JevBench public subset, the assembled backbone and retained candidate heads select an accepted alternative in 146 of 231 examples (63.20\%). No exact normalized-state overlap is found against the checked training states. This result evaluates ranking over the supplied alternatives in that subset. It does not measure an action's physical outcome or the benchmark's sealed-board composite score.

The historical head-validation experiment addresses a separate question: how well the three scorers fit their respective tasks with the paired frozen reference backbone. The selected checkpoint is epoch six of 30, chosen by equal-task macro cross-entropy of 0.7692. Table~\ref{tab:head} shows overall accuracy of 73.58\% across 3,202 examples, but substantial variation between tasks. The largest task, \texttt{noul}, is also the most accurate; the overall sample-weighted score consequently obscures the weaker \texttt{score} task. Equal-task averaging removes the additional weight that the larger task would receive from its sample count. Differences in backbone and evaluation set prevent a direct gain or degradation claim between this component validation and JevBench.

\begin{table}[htbp]
\centering\small
\caption{Historical candidate-head validation with its paired frozen reference backbone. Accuracy follows the recorded criterion allowing tied target alternatives. The overall row is sample-weighted; the equal-task macro cross-entropy used for checkpoint selection is 0.7692.}
\label{tab:head}
\begin{tabular}{@{}lrrr@{}}
\toprule
Task & Examples & Accuracy (\%) & Cross-entropy\\
\midrule
\texttt{choice} & 943 & 72.43 & 0.7916\\
\texttt{noul} & 1,480 & 84.05 & 0.3448\\
\texttt{score} & 779 & 55.07 & 1.1713\\
\midrule
Overall & 3,202 & 73.58 & 0.6774\\
\bottomrule
\end{tabular}
\end{table}

BFCL generation is complete for all 3,641 single-turn cases. The original scorer records 2,871 correct (78.85\%), with no length-limited outputs and one missing final response. Uniform saved-answer rescoring and independent final verification are still pending. The original score is therefore kept separate from an accepted corrected score. Appendix~\ref{app:bfcl} describes the absent-tool request normalization needed for that check. Multi-turn and agentic categories are outside this experiment, so this number is not a complete BFCL v4 overall score.

The two structured-output evaluations serve complementary purposes: BFCL assesses generated function selections and arguments, whereas JevBench assesses ranking over supplied candidates. Together, they characterize the generative and discriminative interfaces available to the application.

\subsection{Inference capacity under growing responses}
\subsubsection{Short-window throughput and sustained cache pressure}
The visual concurrency sweep reaches 8, 16, 32, 64, 128, and 256 actual scheduled requests. Its corresponding 90-second rates are 178.81, 314.81, 490.00, 718.11, 940.81, and 1,020.65 output tokens/s. More active requests improve token production within these windows. However, a subsequent fixed 256-request run reaches 99.62\% occupancy of its 32\,GiB KV cache and accumulates 675 preemptions before termination. The highest measured short-window rate therefore does not establish a stable operating point for long responses.

This behavior is consistent with continued context growth after admission: already active requests consume additional cache even when intake pauses. A high window rate measures productive computation during that window, while sustained completion also depends on whether the accumulated contexts fit. The result motivates the growth-aware admission experiment in Section~\ref{sec:systems-design}, rather than selecting concurrency from the highest short-window rate alone.

\subsubsection{Complete batches and the final response tail}
With 32\,GiB KV cache and a 48\,GiB CUDA allocator cap, a complete 128-request text batch generates 775,019 tokens in 1,659.47 seconds (467.03 tokens/s); an eight-request visual batch generates 28,633 tokens in 615.16 seconds (46.55 tokens/s). Both complete without preemptions, length-limited outputs, or missing final responses, and saved token totals match engine counts. The text batch reaches sampled peak KV occupancy of 67.89\% and CUDA allocation of 41.25\,GiB. These measurements include the tail in which only the longest responses remain active.

\begin{figure}[tb]
\centering
\includegraphics[width=\linewidth]{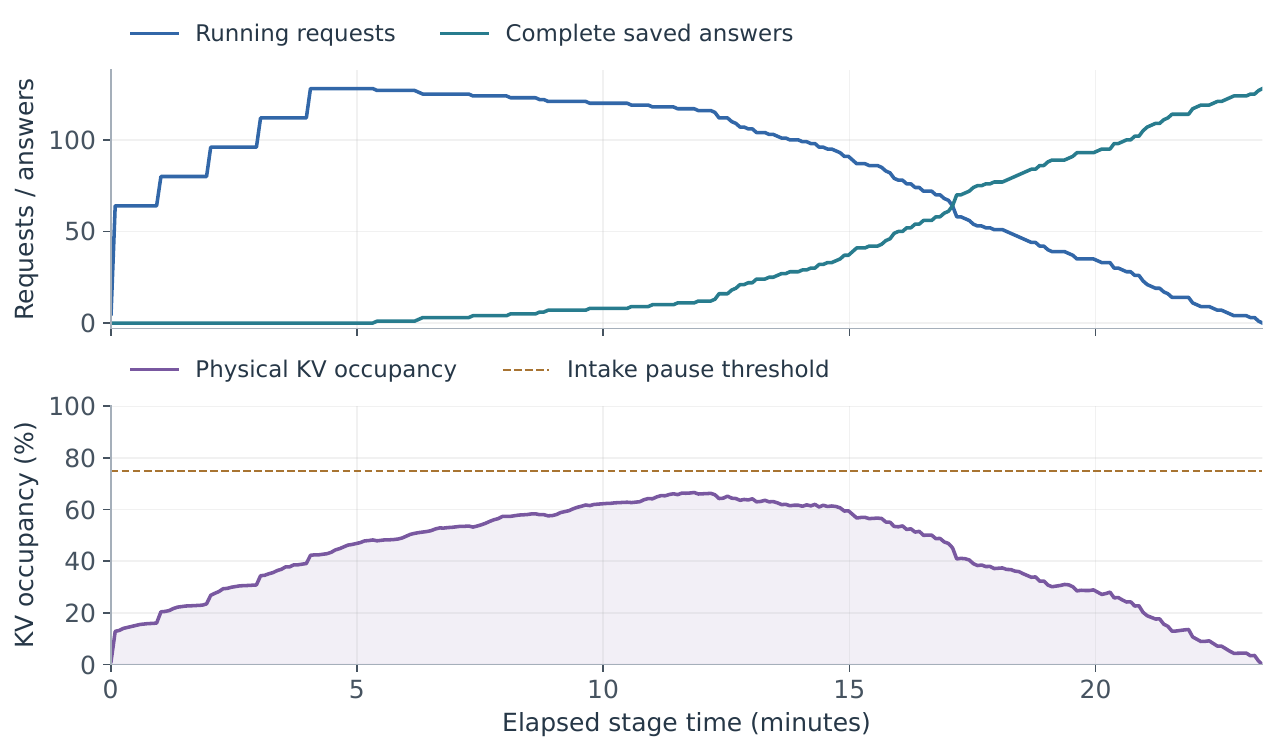}
\caption{An independently captured trace of the initial complete 128-answer MMLU-Pro batch. Top: actual running requests and complete saved answers. Bottom: observed physical KV occupancy; the dashed line marks the intake-pause threshold. All requests finish naturally, and no preemption is recorded. Samples stop before continuous refill resumes. The trace demonstrates this batch's behavior, not long-run stability or steady-state decode throughput.}
\label{fig:cache}
\end{figure}

Under the physical-usage and growth-lease policy, a separate initial batch of 128 new MMLU-Pro answers produces 834,117 saved output tokens in 1,403.27746 seconds, or 594.41 tokens/s. Independent checks match saved tokens to the engine counter difference and find zero preemptions, length-limited outputs, and missing final responses. Previously completed source answers remain unchanged. Sampled peak KV occupancy is 66.63\%. Figure~\ref{fig:cache} shows cache growing while few answers are complete, then draining as long responses finish; both phases contribute to the reported wall time.

This trace establishes completion for the measured batch. Replacement admissions are disabled during its drain, and the examples come from one MMLU-Pro prefix. It therefore does not establish stability under continuous refill or across workloads. Likewise, differences in prompts, cache settings, output lengths, and timing windows prevent interpreting 594.41 tokens/s relative to the preceding rates as an isolated policy speedup.

For the \ThinkingMMMUFreshCount\ fresh MMMU answers, saved output and the drained engine counter both contain \ThinkingMMMUFreshTokens\ tokens. Their generation interval is \ThinkingMMMUSeconds\ seconds, giving \ThinkingMMMURate\ complete-output tokens/s including the final single-request tail. The 455 retained answers contribute to accuracy but add no tokens or time to this later stage. Its two length-limited and four missing-delimiter outputs remain counted. Together, these measurements show why throughput should be accompanied by output-completion checks and a declared workload. Historical short-output and prefix-cache controls are retained in Appendix~\ref{app:historical-performance}.

\subsection{Streaming speech latency and delivery}
\label{sec:speech-results}
Table~\ref{tab:speech-latency} separates the merged model's short synthesis trace from the pre-merge reference runtime's conversational traces. In the merged-model case, the first PCM chunk arrives after 0.390\,s and the final event after 1.305\,s. Four chunks contain 2.48\,s of audio, yielding a request-level RTF of 0.53. This is a warm text-to-speech observation with fixed input text: language-response generation and speech recognition are not part of that request. The script's earlier full-waveform synthesis request takes 5.701\,s for 2.72\,s of audio. Different request formats and execution order prevent treating these two observations as a controlled cold-versus-warm comparison.

\begin{table}[htbp]
\centering\small
\caption{Single-request speech-delivery observations on GB10; one observation per row. Times start before local HTTP submission. The first row measures short fixed-text synthesis with the merged model; the remaining rows use a separate pre-merge reference runtime. PCM arrival is not physical playback. RTF is complete-request time divided by output-audio duration.}
\label{tab:speech-latency}
\begin{tabularx}{\linewidth}{@{}Yrrrr@{}}
\toprule
Runtime and input & First PCM (s) & Complete (s) & Audio (s) & RTF\\
\midrule
Merged; warm fixed-text synthesis & 0.390 & 1.305 & 2.48 & 0.53\\
Reference; text-initiated reply & 1.160 & 16.494 & 19.60 & 0.84\\
Reference; prerecorded audio + image & 4.544 & 29.460 & 29.76 & 0.99\\
\bottomrule
\end{tabularx}
\end{table}

The reference runtime generates a 19.60\,s spoken reply to a text prompt, receiving the first PCM at 1.160\,s and completing the request at 16.494\,s. With prerecorded synthetic speech and an image, first PCM arrives at 4.544\,s and completion at 29.460\,s for 29.76\,s of output audio. First text events occur at 0.119\,s and 3.604\,s, respectively. The multimodal path includes more processing stages, but different inputs and response lengths prevent attributing the latency difference to one stage.

Maximum audio-arrival gaps are 1.018\,s and 1.004\,s for these two reference cases. Replaying their saved arrival traces through the adaptive player produces no buffer underruns. A separate real-model request in a headless Firefox browser records first PCM reception at 1.28\,s and a scheduled audio-context start at 1.31\,s, also with no counted underruns. Scheduled playback and trace replay do not establish physical speaker output or perceptual quality. Although each tabled request has an RTF below one, sustained interaction also depends on startup delay, arrival variability, buffering, and subsequent turns. These single cases support streaming feasibility, not a percentile latency claim or a speed ranking against other models. In particular, the theoretical cold-start first-packet value reported for Qwen3-Omni~\citep{Xu2025Omni} has a different timing definition and is not a matched comparator.

\section{Discussion and Limitations}
\label{sec:limitations}
The experiments support two observations about evaluating a modular system. First, task-level aggregates conceal variation: MMMU subjects, ChartQA question sources, and candidate-head tasks show different error rates. Second, measurement conditions influence what a result can establish. A changed generation protocol cannot isolate a component's contribution, and a short throughput window cannot establish completion under sustained cache growth. Keeping sample selection, generation, scoring, and timing explicit makes these distinctions testable.

\paragraph{Attribution and comparison.}
The pretrained visual-language and speech components supply substantial inherited capability. The published comparisons contextualize the scores, but do not quantify capability retained or lost during assembly. A matched control would hold prompts, image processing, seeds, and checkers fixed while changing only the base or assembled checkpoint. A separate connector study would hold the acoustic inputs and language backbone fixed. These controls are needed to evaluate the contribution of local adaptation; neither the training recipe nor successful component loading supplies that evidence. Our broader speech and candidate-scoring interfaces also require additional stored components, so comparison at language-backbone scale is not a comparison at equal total system size.

\paragraph{Coverage and scoring.}
The long-thinking MMLU-Pro result covers a proportional 3,000-question subset, while BFCL covers only single-turn cases. Independent final MMLU-Pro rescoring and uniform BFCL saved-answer rescoring remain outstanding. IFEval, MATH-500, and ChartQA retain their restricted-protocol measurements. The study provides single-run observations, without repeated-run confidence intervals or same-protocol peer-model comparisons. Public task scores and historical head validation also leave semantic training overlap and performance on unseen application states unresolved.

\paragraph{Speech and interactive behavior.}
The speech traces establish specific delivery behavior, not general speech quality. Recognition error, spoken-content fidelity, speaker similarity, and perceptual naturalness need their own evaluations. The merged-model latency result concerns one warm fixed-text synthesis request, whereas the longer conversational traces precede checkpoint merge and use a separate reference runtime. Neither supplies microphone-to-speaker latency, cold/warm percentiles, or a controlled comparison with an end-to-end speech model. Existing functional checks of reference-voice synthesis and subsequent requests also do not establish speaker similarity or cancellation-response time. Native text and visual benchmark results do not constitute full speech-path acceptance.

\paragraph{Systems and application scope.}
Inference results are limited to one GB10 platform and the declared workloads. The growth leases estimate future demand and do not guarantee a worst-case memory bound; repeated sustained runs are needed to characterize robustness. At the application boundary, a candidate ranking or valid tool call remains an advisory output. Physical execution depends on authorization, current device state, and client validation. No device intervention or clinical study is reported, and no clinical or autonomous-action safety conclusion follows from the benchmark results.

\section{Conclusion}
\model\ combines visual-language reasoning, temporally adapted speech input, reference-conditioned speech output, and supplied-candidate scoring in an independently loadable checkpoint. Its independently rescored MMMU validation score of 74.22\% exceeds the reported validation scores of Gemma 3 4B IT and InternVL3.5-4B without test-time scaling. Its IFEval and MATH-500 scores also exceed the corresponding original Qwen3-4B non-thinking reports. Streaming speech synthesis achieves a first-chunk latency of 390\,ms on GB10, while a complete 128-answer text batch reaches 594.41 output tokens/s without preemptions. These results characterize a system that combines multimodal communication with structured application interfaces. Further work will extend speech-quality evaluation, component studies, and sustained-load measurements. Model weights, code, and redistributable evaluation artifacts are planned for public release, with the model and code under Apache License 2.0 and applicable upstream terms retained.

\section*{Data Availability Statement}
We plan to release the model weights and accompanying code under Apache License 2.0, retaining upstream attribution and applicable third-party terms. This is a release plan; a public repository and release identifier have not yet been assigned. The public evaluation datasets, scoring software, and official comparison sources are identified in the references. Local prediction files, configuration snapshots, and source hashes support the reported measurements. We also plan to provide the evaluation configurations and redistributable result artifacts. Proprietary application data, recordings, and credentials are excluded.

\section*{Ethics Declaration}
Reference-voice conditioning should be used with permission from the speaker and with appropriate disclosure of synthetic speech. This report does not present a human-subject listening study or clinical validation. Candidate scores are advisory model outputs, and physical actions require application-level authorization and validation. No claim of certified medical or autonomous device safety is made.

\section*{Authors and Affiliations}
Zeyuan Li is affiliated with Yujie Intelligence, Hangzhou, China.\\
Correspondence: \mbox{\href{mailto:lizeyuan@drvexo.com}{\texttt{lizeyuan@drvexo.com}}}.

\section*{Conflict of Interest Statement}
The system is developed by Yujie Intelligence. Individual financial and non-financial disclosures remain to be confirmed by the submission authors.

\section*{Funding Acknowledgment}
Funding and grant information are pending author confirmation.

\clearpage
\appendix
\section{Restricted-Protocol Output Diagnostics}
\label{app:protocol-diagnostics}
\begin{figure}[htbp]
\centering
\includegraphics[width=\linewidth]{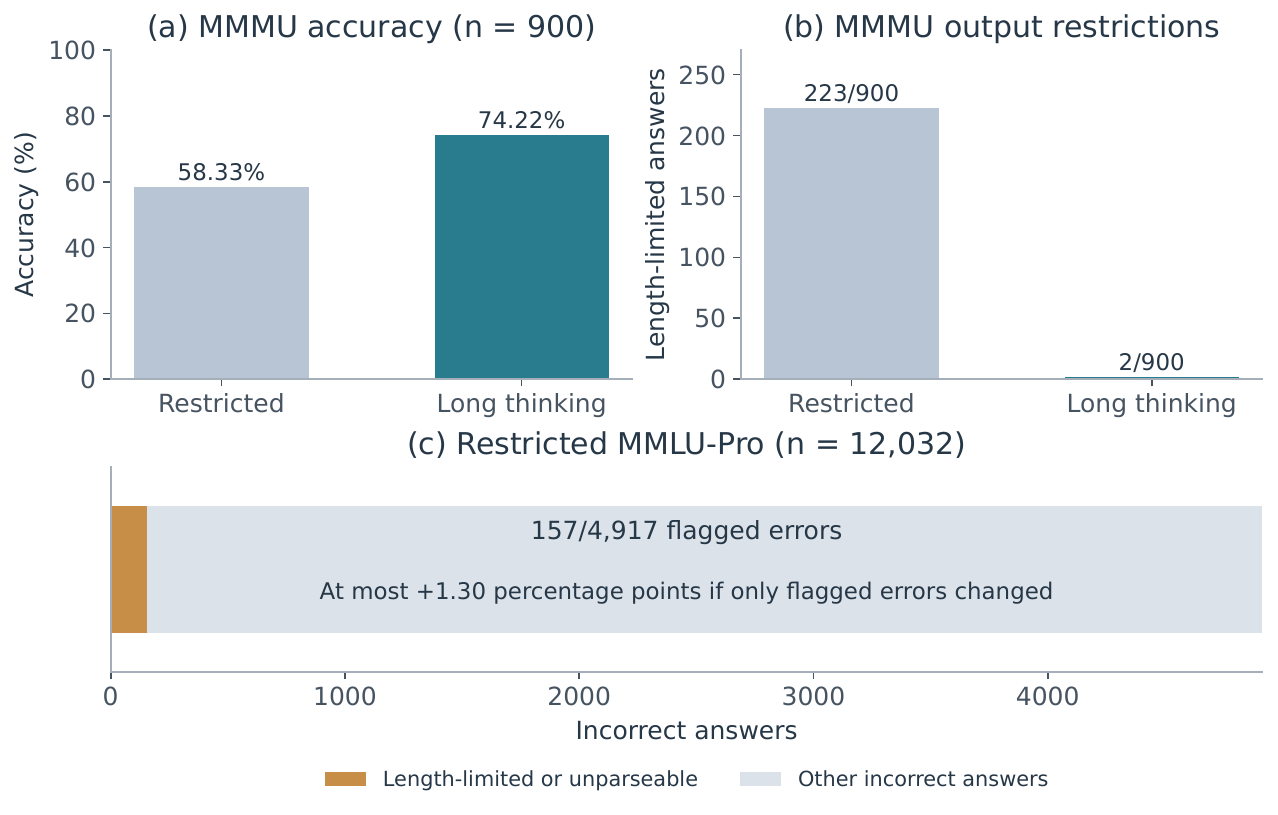}
\caption{Supplementary protocol diagnostics from complete saved outputs. (a) MMMU accuracy under two protocols. (b) Corresponding length-limited output counts. Several settings change jointly. (c) Restricted-protocol MMLU-Pro errors: only 157 of 4,917 carry an observed length-limit or parsing flag. The 1.30-point bound changes only these flagged errors and is not a corrected accuracy or predicted reevaluation gain.}
\label{fig:protocol}
\end{figure}

Figure~\ref{fig:protocol} summarizes output restrictions in the saved answers. Its MMMU panels provide the accuracy and output-limit comparison discussed in Section~\ref{sec:visual-results}. These observations do not isolate the effect of truncation: several settings change jointly, and a length-limited response may already contain a correct answer.

The restricted-protocol MMLU-Pro run contains 160 length-limited outputs in total. Among its errors, only 157 of 4,917 incorrect answers are length-limited or unparseable. Making all 157 correct while holding every other answer fixed would yield at most $157/12{,}032=1.30$ percentage points of improvement. The flagged failures therefore explain only a small part of the observed errors. This bound neither corrects the score nor predicts the long-thinking result; it identifies the limit of an explanation based solely on these two flags. Prompt format, exemplar retention, reasoning behavior, and unflagged answer-extraction errors require separate controls.

\section{Historical Performance Controls}
\label{app:historical-performance}
Table~\ref{tab:fullperf} preserves complete restricted-protocol workload measurements. They use shorter output budgets and less memory than the long-thinking experiments, so the rates must not be ranked as if they measured a common workload.

\begin{table}[htbp]
\centering\small
\caption{Complete restricted-protocol workloads on GB10: at most 128 active sequences, 6\,GiB KV cache, and a 16\,GiB allocator cap. Generation wall time includes interleaved checking and excludes initial loading.}
\label{tab:fullperf}
\begin{tabular}{@{}lrrr@{}}
\toprule
Workload & Output tokens & Time (s) & Tokens/s\\
\midrule
IFEval, 541 prompts & 169,134 & 275.43 & 614.06\\
MATH-500, 500 problems & 609,240 & 1,041.74 & 584.83\\
\bottomrule
\end{tabular}
\end{table}

On the same 128 IFEval prompts, 64 active sequences with 4\,GiB KV yield 345.8 tokens/s and 128 with 6\,GiB KV yield 349.6 tokens/s. Wall time falls from 114.19 to 113.31 seconds, approximately 0.78\%. Both concurrency and cache capacity change, so this comparison does not isolate either factor.

On a fixed 128-question MMLU-Pro probe, disabling and enabling prefix caching yields 102.4 and 245.7 tokens/s, respectively, with wall times of 123.48 and 37.72 seconds. The cache-enabled log reports an 85.5\% hit rate. Output tokens also change from 12,646 to 9,270, four answer choices change, and 102 outputs are text-identical. Probe correctness changes from 105 to 106/128, and length-limited outputs fall from two to zero. This is an observed workload difference with altered output lengths, rather than an identical-output cache-only speedup. No representative MMLU-Pro accuracy or repeated-run uncertainty is inferred from this probe.

\section{BFCL Scoring Preparation}
\label{app:bfcl}
\paragraph{Absent-tool request normalization.}
A CPU preflight before BFCL generation identifies four examples with empty tool lists among the 3,641 inputs. The installed parser request wrapper rejects an empty \texttt{tools} array together with automatic tool selection before parsing a response. A separate adapter omits both request fields only when no tools are supplied. CPU checks preserve the request fields for nonempty-tool representatives of all 13 categories and preserve positive and negative official-checker controls; a missing thinking delimiter remains invalid.

All 3,641 answers have been generated, but uniform saved-answer rescoring and independent verification remain pending; no corrected BFCL accuracy is available. The original run and its 78.85\% score are preserved. Final acceptance requires unchanged answer text, IDs, order, token counts and stop reasons; any scoring change must be restricted to the four absent-tool cases with the recorded request-schema exception. The verification must compare source and corrected-output hashes and enumerate the changed cases. It must not regenerate or select responses again.

\section{Reproducibility Record}
\label{app:reproducibility}
The checkpoint-size accounting reads only safetensors metadata headers and sums the product of each stored tensor's dimensions. It identifies 6,970,568,132 scalar elements across 2,131 tensors. This includes packaged speaker state and other buffers, and is reported as a storage count rather than a count of unique trainable parameters. Mixed BF16 and FP32 storage prevents inferring this count by dividing checkpoint bytes by a single element size.

The evidence snapshots identify the package index, input and prediction hashes, evaluation runner, precision-restoration worker, scoring helpers, and serving configuration. The long-response supplement preserves the complete-batch acceptance record, independent token and score audits, initial-drain checkpoint, and 278 telemetry records preceding refill. MMMU throughput is bound to the stage that generated its 445 fresh answers; later reuse of all 900 completed answers contributes no new generation time or token rate.

A separate completed-run supplement preserves the MMLU-Pro subset manifest and the MMLU-Pro and BFCL result records with their prediction hashes. It records 3,000 unique MMLU-Pro answers, including 624 retained responses, and 3,641 unique BFCL answers. These files support the original-run scores in Table~\ref{tab:task-details}; they do not substitute for the pending independent final rescoring or BFCL normalization check. The subset manifest fixes selection by ID and subject without consulting generated answers or correctness.

\begin{table}[htbp]
\centering\small
\caption{Complete DRVEXO task counts under the protocols in Section~\ref{sec:protocols}. [a] denotes a completed-run score awaiting independent final rescoring. [b] denotes the original BFCL single-turn score, awaiting uniform saved-answer rescoring and independent verification. The MMLU-Pro subset and full-set results use different evaluated samples.}
\label{tab:task-details}
\begin{tabularx}{\linewidth}{@{}lYrr@{}}
\toprule
Task & Protocol / metric & Correct / total & Score (\%)\\
\midrule
IFEval~\citep{Zhou2023IFEval} & Restricted; strict prompt & 445 / 541 & 82.26\\
       & Restricted; strict instruction & 730 / 834 & 87.53\\
MATH-500~\citep{H42024Math} & Restricted; answer accuracy & 447 / 500 & 89.40\\
MMLU-Pro~\citep{Wang2024MMLUPro} & Restricted; full set & 7,115 / 12,032 & 59.13\\
         & Long thinking; stratified subset & \SubsetMMLUCorrect\ / 3,000 & \SubsetMMLUScore\ [a]\\
MMMU~\citep{Yue2024MMMU} & Restricted; public validation & 525 / 900 & 58.33\\
     & Long thinking; public validation & \ThinkingMMMUCorrect\ / 900 & \ThinkingMMMUScore\\
ChartQA~\citep{Masry2022ChartQA} & Restricted; public validation & 1,567 / 1,920 & 81.61\\
JevBench~\citep{BenchmarkHeaven2026JevBench} & Reference backend; public subset & 146 / 231 & 63.20\\
BFCL~\citep{Patil2025BFCL} & Long thinking; single-turn cases & 2,871 / 3,641 & 78.85 [b]\\
\bottomrule
\end{tabularx}
\end{table}

The 594.41 tokens/s batch rate is recomputed from 834,117 tokens and the stored elapsed time of 1,403.27746 seconds. The checkpoint's convenience rate used a clock sampled 0.273 seconds later after source-binding checks; the independent audit brackets that later sampling time with telemetry. The reported value uses the explicit elapsed interval, preserving the frozen checkpoint and the distinction between a complete batch and subsequent continuous refill.

Table~\ref{tab:repro} lists the restricted-protocol generation settings; the long-thinking protocol appears in the experiments section. Local filesystem and service identifiers are omitted from the manuscript because they are not portable reproduction instructions. Public artifact hosting and a complete clean-environment reproduction procedure remain publication tasks.

\begin{table}[htbp]
\centering\small
\caption{Restricted-protocol settings for the completed text scores. The long-thinking protocol is specified in Section~\ref{sec:protocols}.}
\label{tab:repro}
\begin{tabularx}{\linewidth}{@{}lY@{}}
\toprule
Setting & Value\\
\midrule
Checkpoint & Complete BF16 release; original FP32 parameters preserved\\
Hardware / text engine & NVIDIA GB10 / vLLM 0.33\\
Decoding / context & Greedy / 8,192 tokens\\
Active sequences / KV cache & At most 128 / 6\,GiB\\
Prefill budget / prefix caching & 2,048 tokens / enabled\\
IFEval output budget & 2,048 new tokens\\
MATH-500 output budget & 4,096 new tokens\\
MMLU-Pro prompt / output & Raw completion; up to five same-subject exemplars under the 2,048-input-token rule / 2,048 new tokens\\
Memory constraints & 16\,GiB CUDA allocator cap; 90\,GiB process cap\\
\bottomrule
\end{tabularx}
\end{table}

\bibliographystyle{plainnat}
\bibliography{references}

\begin{thebibliography}{23}
\providecommand{\natexlab}[1]{#1}
\providecommand{\url}[1]{\texttt{#1}}
\expandafter\ifx\csname urlstyle\endcsname\relax
  \providecommand{\doi}[1]{doi: #1}\else
  \providecommand{\doi}{doi: \begingroup \urlstyle{rm}\Url}\fi

\bibitem[Bai et~al.(2025)Bai, Cai, Chen, et~al.]{Bai2025QwenVL}
Shuai Bai, Yuxuan Cai, Ruizhe Chen, et~al.
\newblock {Qwen3-VL Technical Report}.
\newblock \emph{arXiv preprint arXiv:2511.21631}, 2025.
\newblock \doi{10.48550/arXiv.2511.21631}.
\newblock URL \url{https://arxiv.org/abs/2511.21631v2}.
\newblock Version 2, Table 4.

\bibitem[{Benchmark Heaven}(2026)]{BenchmarkHeaven2026JevBench}
{Benchmark Heaven}.
\newblock {JevBench}: Benchmark for {AI} decision models.
\newblock Public benchmark software and methodology, 2026.
\newblock URL \url{https://github.com/fstandhartinger/jevbench}.
\newblock Public subset; distinct from the sealed leaderboard. Accessed
  November 5, 2026.

\bibitem[{Gemma Team}(2025)]{Gemma2025Report}
{Gemma Team}.
\newblock Gemma 3 technical report.
\newblock \emph{arXiv preprint arXiv:2503.19786}, 2025.
\newblock \doi{10.48550/arXiv.2503.19786}.
\newblock URL \url{https://arxiv.org/abs/2503.19786v1}.
\newblock Version 1, Table 16.

\bibitem[Hendrycks et~al.(2021)Hendrycks, Burns, Kadavath,
  et~al.]{Hendrycks2021Math}
Dan Hendrycks, Collin Burns, Saurav Kadavath, et~al.
\newblock Measuring mathematical problem solving with the {MATH} dataset.
\newblock \emph{arXiv preprint arXiv:2103.03874}, 2021.
\newblock \doi{10.48550/arXiv.2103.03874}.
\newblock URL \url{https://arxiv.org/abs/2103.03874}.

\bibitem[{Hugging Face}(2025)]{HuggingFace2025MathVerify}
{Hugging Face}.
\newblock {Math-Verify}: A mathematical answer verification library.
\newblock Software repository, 2025.
\newblock URL \url{https://github.com/huggingface/Math-Verify}.
\newblock Evaluation uses version 0.9.0; accessed November 5, 2026.

\bibitem[{HuggingFaceH4}(2024)]{H42024Math}
{HuggingFaceH4}.
\newblock {MATH-500}.
\newblock Hugging Face dataset repository, 2024.
\newblock URL \url{https://huggingface.co/datasets/HuggingFaceH4/MATH-500}.
\newblock Dataset card and evaluation split; accessed November 5, 2026.

\bibitem[Kwon et~al.(2023)Kwon, Li, Zhuang, Sheng, Zheng, Yu, Gonzalez, Zhang,
  and Stoica]{Kwon2023PagedAttention}
Woosuk Kwon, Zhuohan Li, Siyuan Zhuang, Ying Sheng, Lianmin Zheng, Cody~Hao Yu,
  Joseph~E. Gonzalez, Hao Zhang, and Ion Stoica.
\newblock Efficient memory management for large language model serving with
  {PagedAttention}.
\newblock \emph{arXiv preprint arXiv:2309.06180}, 2023.
\newblock \doi{10.48550/arXiv.2309.06180}.
\newblock URL \url{https://arxiv.org/abs/2309.06180}.

\bibitem[Masry et~al.(2022)Masry, Long, Tan, Joty, and Hoque]{Masry2022ChartQA}
Ahmed Masry, Do~Xuan Long, Jia~Qing Tan, Shafiq Joty, and Enamul Hoque.
\newblock {ChartQA}: A benchmark for question answering about charts with
  visual and logical reasoning.
\newblock In \emph{Findings of the Association for Computational Linguistics:
  ACL 2022}, pages 2263--2279, 2022.
\newblock \doi{10.18653/v1/2022.findings-acl.177}.
\newblock URL \url{https://aclanthology.org/2022.findings-acl.177/}.

\bibitem[{OpenBMB}(2025)]{OpenBMB2025MiniCPMV4}
{OpenBMB}.
\newblock {MiniCPM-V 4.0} model card.
\newblock Hugging Face model repository, 2025.
\newblock URL \url{https://huggingface.co/openbmb/MiniCPM-V-4}.
\newblock Revision 6f69a8235885be89608f3a71ffd40695b379e1e7; accessed November
  7, 2026.

\bibitem[Patil et~al.(2025)Patil, Mao, Yan, Ji, Suresh, Stoica, and
  Gonzalez]{Patil2025BFCL}
Shishir~G Patil, Huanzhi Mao, Fanjia Yan, Charlie Cheng-Jie Ji, Vishnu Suresh,
  Ion Stoica, and Joseph~E. Gonzalez.
\newblock The berkeley function calling leaderboard ({BFCL}): From tool use to
  agentic evaluation of large language models.
\newblock In \emph{Proceedings of the 42nd International Conference on Machine
  Learning}, volume 267 of \emph{Proceedings of Machine Learning Research},
  pages 48371--48392. PMLR, 2025.
\newblock URL \url{https://proceedings.mlr.press/v267/patil25a.html}.

\bibitem[{Qwen ASR Team}(2026)]{Qwen2026ASR}
{Qwen ASR Team}.
\newblock {Qwen3-ASR Technical Report}.
\newblock \emph{arXiv preprint arXiv:2601.21337}, 2026.
\newblock \doi{10.48550/arXiv.2601.21337}.
\newblock URL \url{https://arxiv.org/abs/2601.21337}.

\bibitem[{Qwen Team}(2025{\natexlab{a}})]{Qwen2025Instruct2507Card}
{Qwen Team}.
\newblock {Qwen3-4B-Instruct-2507} model card.
\newblock Hugging Face model repository, 2025{\natexlab{a}}.
\newblock URL \url{https://huggingface.co/Qwen/Qwen3-4B-Instruct-2507}.
\newblock Revision cdbee75f17c01a7cc42f958dc650907174af0554; original Qwen3-4B
  Non-Thinking comparator column; accessed November 7, 2026.

\bibitem[{Qwen Team}(2025{\natexlab{b}})]{Qwen2025Thinking2507Card}
{Qwen Team}.
\newblock {Qwen3-4B-Thinking-2507} model card.
\newblock Hugging Face model repository, 2025{\natexlab{b}}.
\newblock URL \url{https://huggingface.co/Qwen/Qwen3-4B-Thinking-2507}.
\newblock Revision 768f209d9ea81521153ed38c47d515654e938aea; original Qwen3-4B
  Thinking comparator column; accessed November 7, 2026.

\bibitem[{Qwen Team}(2026{\natexlab{a}})]{Qwen2026Model}
{Qwen Team}.
\newblock {Qwen3.5-4B} model card.
\newblock Hugging Face model repository, 2026{\natexlab{a}}.
\newblock URL \url{https://huggingface.co/Qwen/Qwen3.5-4B}.
\newblock Benchmark table checked at revision
  851bf6e806efd8d0a36b00ddf55e13ccb7b8cd0a; accessed November 7, 2026.

\bibitem[{Qwen Team}(2026{\natexlab{b}})]{Qwen2026TTS}
{Qwen Team}.
\newblock {Qwen3-TTS-12Hz-1.7B-Base} model card.
\newblock Hugging Face model repository, 2026{\natexlab{b}}.
\newblock URL \url{https://huggingface.co/Qwen/Qwen3-TTS-12Hz-1.7B-Base}.
\newblock Accessed November 5, 2026.

\bibitem[{Qwen Team}(2026{\natexlab{c}})]{Qwen2026TTSPaper}
{Qwen Team}.
\newblock {Qwen3-TTS Technical Report}.
\newblock \emph{arXiv preprint arXiv:2601.15621}, 2026{\natexlab{c}}.
\newblock \doi{10.48550/arXiv.2601.15621}.
\newblock URL \url{https://arxiv.org/abs/2601.15621}.

\bibitem[Wang et~al.(2025)Wang, Gao, Gu, et~al.]{Wang2025InternVL}
Weiyun Wang, Zhangwei Gao, Lixin Gu, et~al.
\newblock {InternVL3.5}: Advancing open-source multimodal models in
  versatility, reasoning, and efficiency.
\newblock \emph{arXiv preprint arXiv:2508.18265}, 2025.
\newblock \doi{10.48550/arXiv.2508.18265}.
\newblock URL \url{https://arxiv.org/abs/2508.18265v2}.
\newblock Version 2, Tables 3 and 4; no test-time scaling.

\bibitem[Wang et~al.(2024)Wang, Ma, Zhang, et~al.]{Wang2024MMLUPro}
Yubo Wang, Xueguang Ma, Ge~Zhang, et~al.
\newblock {MMLU-Pro}: A more robust and challenging multi-task language
  understanding benchmark.
\newblock \emph{arXiv preprint arXiv:2406.01574}, 2024.
\newblock \doi{10.48550/arXiv.2406.01574}.
\newblock URL \url{https://arxiv.org/abs/2406.01574}.

\bibitem[Xu et~al.(2025{\natexlab{a}})Xu, Peng, Awadalla,
  et~al.]{Xu2025Phi4Mini}
Haoran Xu, Baolin Peng, Hany Awadalla, et~al.
\newblock {Phi-4-Mini-Reasoning}: Exploring the limits of small reasoning
  language models in math.
\newblock \emph{arXiv preprint arXiv:2504.21233}, 2025{\natexlab{a}}.
\newblock \doi{10.48550/arXiv.2504.21233}.
\newblock URL \url{https://arxiv.org/abs/2504.21233v1}.
\newblock Version 1, Table 1.

\bibitem[Xu et~al.(2025{\natexlab{b}})Xu, Guo, Hu, et~al.]{Xu2025Omni}
Jin Xu, Zhifang Guo, Hangrui Hu, et~al.
\newblock {Qwen3-Omni Technical Report}.
\newblock \emph{arXiv preprint arXiv:2509.17765}, 2025{\natexlab{b}}.
\newblock \doi{10.48550/arXiv.2509.17765}.
\newblock URL \url{https://arxiv.org/abs/2509.17765}.

\bibitem[Yang et~al.(2025)Yang, Li, Yang, et~al.]{Yang2025Qwen3}
An~Yang, Anfeng Li, Baosong Yang, et~al.
\newblock {Qwen3 Technical Report}.
\newblock \emph{arXiv preprint arXiv:2505.09388}, 2025.
\newblock \doi{10.48550/arXiv.2505.09388}.
\newblock URL \url{https://arxiv.org/abs/2505.09388v1}.
\newblock Version 1, Section 4.6 and Tables 17 and 18.

\bibitem[Yue et~al.(2023)Yue, Ni, Zhang, et~al.]{Yue2024MMMU}
Xiang Yue, Yuansheng Ni, Kai Zhang, et~al.
\newblock {MMMU}: A massive multi-discipline multimodal understanding and
  reasoning benchmark for expert {AGI}.
\newblock \emph{arXiv preprint arXiv:2311.16502}, 2023.
\newblock \doi{10.48550/arXiv.2311.16502}.
\newblock URL \url{https://arxiv.org/abs/2311.16502}.

\bibitem[Zhou et~al.(2023)Zhou, Lu, Mishra, et~al.]{Zhou2023IFEval}
Jeffrey Zhou, Tianjian Lu, Swaroop Mishra, et~al.
\newblock Instruction-following evaluation for large language models.
\newblock \emph{arXiv preprint arXiv:2311.07911}, 2023.
\newblock \doi{10.48550/arXiv.2311.07911}.
\newblock URL \url{https://arxiv.org/abs/2311.07911}.

\end{thebibliography}

\end{document}